\documentclass[runningheads]{llncs}
\usepackage{graphicx}
\usepackage{soul}
\usepackage{caption}
\usepackage{subfig}
\usepackage{blindtext}
\usepackage{tikz}
\usepackage{comment}
\usepackage{amsmath,amssymb} % define this before the line numbering.
\usepackage{color}
\usepackage{booktabs}
\usepackage{multirow}
\usepackage{pifont}% http://ctan.org/pkg/pifont
\usepackage{bm}

\newcommand{\eg}{\textit{e.g.}}
\newcommand{\ie}{\textit{i.e.}}
\newcommand{\shortmodelname}{TAGS}

\usepackage[accsupp]{axessibility}  % Improves PDF readability for those with disabilities.

\usepackage{color}
\usepackage[pagebackref=true,breaklinks=true,colorlinks]{hyperref}

\begin{document}
% \linenumbers
\pagestyle{headings}
\mainmatter
\def\ECCVSubNumber{100}  % Insert your submission number here

\title{Proposal-Free Temporal Action Detection via Global Segmentation Mask Learning} % Replace with your title

%\begin{comment}
\titlerunning{ECCV-22 submission ID 977} 
\authorrunning{ECCV-22 submission ID 977} 
\author{Anonymous ECCV submission}
\institute{Paper ID 977}
%\end{comment}
%******************

% % CAMERA READY SUBMISSION
% \begin{comment}
\titlerunning{Proposal-Free TAD via Global Segmentation Masking}
\author{Sauradip Nag\inst{1,2} \and
Xiatian Zhu\inst{1,3} \and
Yi-Zhe Song\inst{1,2} \and
Tao Xiang\inst{1,2}}
\authorrunning{Nag et al.}
\institute{CVSSP, University of Surrey, UK \and
iFlyTek-Surrey Joint Research Centre on Artificial Intelligence, UK \and
Surrey Institute for People-Centred Artificial Intelligence, University of Surrey, UK\\
\email{\{s.nag,xiatian.zhu,y.song,t.xiang\}@surrey.ac.uk}}
% \end{comment}
% %******************
\maketitle

\begin{abstract}
Existing temporal action detection (TAD) methods 
rely on generating an overwhelmingly large number of proposals per video. This leads to  complex model designs due to proposal generation and/or per-proposal action instance evaluation and the resultant high computational cost. In this work, for the first time, we propose a  {\em proposal-free} \textcolor{black}{{\em \underline{T}emporal  \underline{A}ction detection model via \underline{G}lobal \underline{S}egmentation mask} (\shortmodelname)}.  Our core idea is to learn a  {\em global segmentation mask} of each action instance jointly at the full video length.  The {\shortmodelname} model differs significantly from the conventional proposal-based methods by focusing on global temporal representation learning to directly detect local start and end points of action instances without proposals. Further, by modeling TAD holistically rather than locally at the individual proposal level,
{\shortmodelname} needs a much simpler model architecture with lower computational cost.
Extensive experiments show that despite its simpler design, {\shortmodelname}  outperforms existing TAD methods, achieving new state-of-the-art performance on two benchmarks. Importantly, it is  $\sim{20\times}$ faster to train and  $\sim{1.6\times}$ more efficient for inference. Our PyTorch implementation of {\shortmodelname} is available at \href{https://github.com/sauradip/TAGS}{https://github.com/sauradip/TAGS}.
\end{abstract}

\section{Introduction}
\label{sec:intro}
\begin{figure}[!t]
\begin{center}
  \includegraphics[scale = 0.170]{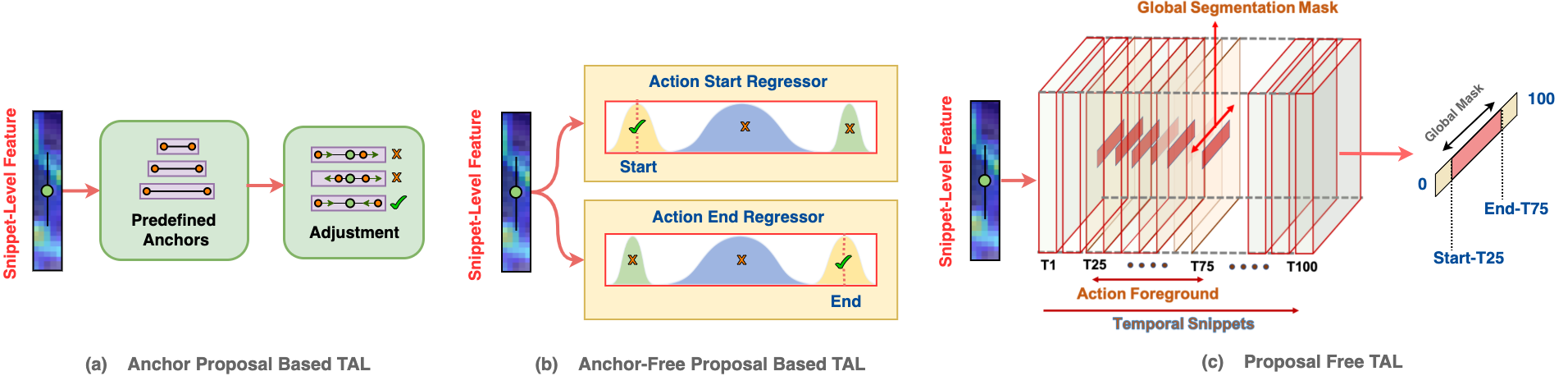}
\end{center}
\caption{%Architecture comparison:
All existing TAD methods,  no matter whether (a) anchor-based or (b) anchor-free,  all
need to generate action proposals.
Instead, (c) our {\em global segmentation mask model} ({\shortmodelname}) is proposal-free.
}
% %\vspace{-.20in}
\label{fig:design}
\end{figure}

Temporal action detection (TAD) aims to identify the temporal interval (\ie, the start and end points) and the class label of all action instances in an untrimmed video \cite{idrees2017thumos,caba2015activitynet}. 
All existing TAD methods 
rely on {\em proposal generation} by either 
regressing predefined anchor boxes 
\cite{xu2017r,chao2018rethinking,gao2017turn,long2019gaussian}
(Fig.~\ref{fig:design}(a))
or 
directly predicting the start and end times of proposals
\cite{lin2019bmn,buch2017sst,lin2018bsn,xu2020g,nag2021few,xu2021boundary,xu2021low}
(Fig.~\ref{fig:design}(b)).
Centered around proposals, existing TAD methods essentially take a local view of the video and focus on each individual proposal for action instance temporal refinement and classification. Such an approach thus
suffers from
several fundamental limitations:
{(1)} An excessive (sometimes exhaustive) number of proposals are usually required
for good performance.
For example, BMN \cite{lin2019bmn} generates $\sim 5000$ proposals per video
by exhaustively pairing start and end points predicted. Generating and evaluating such a large number of proposals means high computational costs for both training and inference.
{(2)}  Once the proposals are generated, the subsequent modeling is local to each individual proposal. Missing global context over the whole video can lead to sub-optimal detection. 

In this work, for the first time, we address these limitations by proposing a proposal-free  TAD model. Our model, termed {\shortmodelname},  learns  a {\em global segmentation mask} of action instances at the full video length (Fig.~\ref{fig:design}(c)). By modeling TAD globally rather than locally, {\shortmodelname} not only removes the need for proposal generation, and the associated design and computational complexity, it is also more effective.  
Concretely, 
instead of predicting the start/end points of each action instance,
{\shortmodelname} learns to predict an action segmentation mask of an entire video. Such a mask represents the {\em global} temporal structure of all action instances in the video; {\shortmodelname} is thus intrinsically global context-aware. 

Taking a proposal-free approach to TAD, our {\shortmodelname} has a simpler model architecture design than existing methods. Specifically, it takes each local snippet (\ie, a short sequence of consecutive frames of a video) as a predictive unit. That is, taking as input a snippet feature representation for a given video,
{\shortmodelname} directly outputs the target action segmentation mask as well as class label concurrently.
To facilitate global context modeling,
we leverage self-attention \cite{vaswani2017attention} 
for capturing necessary video-level inter-snippet relationship. Once the mask is generated, simple foreground segment classification follows to produce the final TAD result. 
\textcolor{black}{To facilitate global segmentation mask learning, we further introduce a novel boundary focused loss that pays more attention to temporal boundary regions, and leverage mask predictive redundancy and inter-branch consistency for
prediction enhancement.} 
During inference, once the masks and class labels are predicted, 
top-scoring segments \textcolor{black}{with refined boundary} can then be selected via non-maximal suppression (NMS)
to produce the final TAD result.

We make the following {\bf contributions}.
{\bf (I)} We present a novel proposal-free TAD model based on {\em global segmentation mask} ({\shortmodelname}) learning. To the best of our knowledge, this is the first model that eliminates the need for proposal generation/evaluation. As a result, it has a much simpler model design with a lower computational cost than existing alternatives. 
{\bf (II)} 
We improve TAD feature representation learning with global 
temporal context using self-attention leading to context-aware TAD. 
{{\bf (III)} To enhance the learning of temporal boundary, {\color{black} we propose a novel boundary focused loss function, along with mask predictive redundancy and inter-branch consistency.}}
{\bf (IV)}
Extensive experiments show that %without bells and whistles 
the proposed {\shortmodelname} method yields new state-of-the-art performance on two TAD datasets (ActivityNet-v1.3 and THUMOS'14).
Importantly, our method is also significantly more efficient in both training/inference. For instance, it is
 {20/1.6}$\times$ faster than G-TAD \cite{xu2020g} in training and inference respectively.

% \vspace{-0.10in}
\section{Related Works}
\label{sec:related}
Although all existing TAD methods use proposals, they differ in how the proposals are generated.

% % % \vspace{0.1cm}
\noindent{\bf Anchor-based proposal learning methods }
These methods generate proposal based on a pre-determined set of anchors. Inspired by object detection in static images \cite{ren2016faster},
R-C3D \cite{xu2017r} proposes to use anchor boxes.
It follows the structure of {proposal generation and classification} in design.
With similar model design, TURN \cite{gao2017turn} aggregates local features to represent snippet-level features, which are then used for temporal boundary regression and classification.
Later, GTAN \cite{long2019gaussian}
improves the proposal feature pooling procedure with a learnable Gaussian kernel for weighted averaging.
PBR-Net \cite{liu2020progressive} improves the detection performance using a pyramidal anchor based detection and fine-grained refinement using frame-level features.
G-TAD \cite{xu2020g}
learns semantic and temporal context via graph convolutional networks for better proposal generation. 
\textcolor{black}{
MUSES \cite{liu2021multi} further improves the performance by handling intra-instance variations caused by shot change.
VSGN \cite{zhao2021video} focuses on short-action detection in a cross-scale multi-level pyramidal architecture.}
Note that these anchor boxes are often exhaustively generated so are high in number.

\noindent{\bf Anchor-free proposal learning methods }
Instead of using pre-designed and fixed anchor boxes, these methods directly learn to predict temporal proposals (\ie, start and end times/points) 
\cite{zhao2017temporal,lin2018bsn,lin2019bmn}.
For example, SSN \cite{zhao2017temporal} decomposes an action instance into three stages (starting, course, and ending)
and employs structured temporal pyramid pooling
to generate proposals.
BSN \cite{lin2018bsn} predicts the start, end and actionness at each temporal location and generates proposals using locations with high start and end probabilities.
Later, BMN \cite{lin2019bmn}
additionally generates a boundary-matching confidence map to improve proposal generation. 
\textcolor{black}{BSN++ \cite{su2020bsn++} further extends BMN with a complementary boundary generator to capture rich context.
CSA \cite{sridhar2021class} enriches the proposal temporal context via attention transfer. 
Recently, ContextLoc \cite{zhu2021enriching} further pushes the boundaries by adapting global context at proposal-level and handling the context-aware inter-proposal relations.}
While no pre-defined anchor boxes are required, these methods often have to exhaustively
pair all possible locations predicted with high scores.
So both anchor-based and anchor-free TAD methods
have a large quantity of temporal proposals to evaluate. This results in complex model design, high computational cost and lack of global context modeling. 
Our {\shortmodelname} is designed to address all these limitations by being proposal-free.

% % % \vspace{0.1cm}
\noindent{\bf Self-attention }
Our snippet representation is learned based on self-attention, which has been firstly introduced in Transformers for natural language processing tasks~\cite{vaswani2017attention}.
In computer vision, non-local neural networks~\cite{wang2018non} apply the core self-attention block from transformers for context modeling and feature learning.
State-of-the-art performance has been achieved in classification~\cite{dosovitskiy2020image}, self-supervised learning~\cite{chen2020generative}, semantic segmentation~\cite{zhang2020dynamic,zheng2020rethinking}, object detection~\cite{carion2020end,yin2020disentangled,zhu2020deformable}, 
few-shot action recognition \cite{perrett2021temporal,zhu2021few},
and object tracking \cite{chen2021transformer}
by using such an attention model. \textcolor{black}{Several recent works \cite{tan2021relaxed,wang2021oadtr,qing2021temporal,nag2022pclfm,nag2022pftm,nag2021few,nag2021temporal} also use 
Transformers for TAD.
They focus on either temporal proposal generation \cite{tan2021relaxed} or refinement \cite{qing2021temporal}.
}
In this paper, we demonstrate the effectiveness of self-attention in a novel proposal-free TAD architecture.

\begin{figure*}[!t]
\begin{center}
  \includegraphics[scale=0.170]{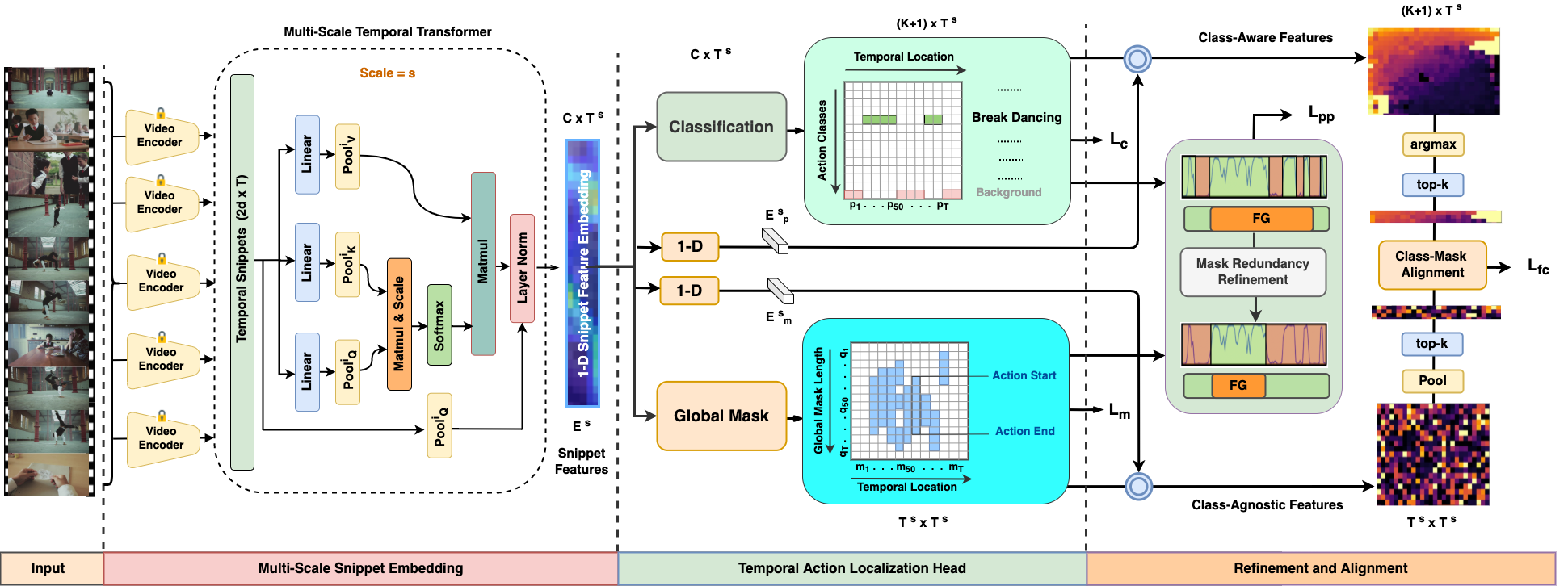}
\end{center}
% % % \vspace{-.2in}
\caption{\textbf{Architecture of our proposal-free {\em \underline{T}emporal  \underline{A}ction detection model via \underline{G}lobal \underline{S}egmentation mask} (\shortmodelname).} 
Given an untrimmed video, {\shortmodelname} first extracts a sequence of $T$ snippet features with a pre-trained video encoder (\eg, I3D \cite{carreira2017quo}), and conducts self-attentive learning {\color{black}at multiple temporal scales $s$} to obtain snippet embedding with global context.
Subsequently, with each snippet embedding, 
{\shortmodelname} classifies different actions (output $\bm{P}^{s} \in \mathbb{R}^{(K+1)\times T^{s}}$ with $K$ the action class number) and 
predicts full-video-long foreground mask (output $\bm{M}^{s} \in \mathbb{R}^{T^{s}\times T^{s}}$) concurrently 
in a two-branch design.
During training, {\shortmodelname} minimizes the difference of
class and mask predictions against the ground-truth.
For more accurate localization, 
an efficient boundary refinement strategy is further introduced,
along with mask predictive redundancy and classification-mask consistency regularization.
During inference, {\shortmodelname} selects top scoring snippets from the classification output $\bm{P}$, and then thresholds the corresponding foreground masks in $\bm{M}$ at each scale and then aggregates them to yield action instance candidates. 
Finally, softNMS is applied to remove redundant candidates. 
}
% \vspace{-.15in}
\label{fig:network}
\end{figure*}

\section{Proposal-Free Global Segmentation Mask}
\label{sec:method}
% \section{Method}
Our {\em global segmentation mask} ({\shortmodelname}) model
takes as input an untrimmed video
$V$ with a variable number of frames.
Video frames are pre-processed by a feature encoder 
(\eg, a Kinetics pre-trained I3D network \cite{carreira2017quo})
into a sequence of localized snippets following the standard practice \cite{lin2019bmn}.
To train the model, we collect a set of labeled 
video training set $\mathcal{D}^{train} = \{V_i, \Psi_i\}$. Each video $V_i$ is labeled with temporal segmentation 
$\Psi_i = \{(\psi_j, \xi_j, y_j)\}_{j=1}^{M_i}$
where $\psi_{j}$/$\xi_{j}$ denote 
the start/end time, $y_j$ is the action category,
and $M_i$ is the action instance number. 

\noindent{\bf Architecture}
As depicted in Fig.~\ref{fig:network}, a {\shortmodelname} model
has two key components:
(1) a self-attentive snippet embedding module
that learns feature representations with global temporal context (Sec.~\ref{sec:model_embedding}),
and
(2) a temporal action detection head with two branches
for per-snippet multi-class action classification 
and binary-class global segmentation mask inference, respectively
 (Sec.~\ref{sec:model_head}).
 
\subsection{Self-attentive multi-scale snippet embedding}
\label{sec:model_embedding}
Given a varying length untrimmed video $V$,
following the standard practice \cite{xu2020g,lin2019bmn} we first sample $T$ equidistantly distributed temporal snippets (points) over the entire length and \textcolor{black}{use a Kinetics pre-trained video encoder 
(\eg, a two-stream model \cite{wang2016temporal}) to extract 
RGB $X_{r} \in \mathbb{R}^{d \times T}$ and optical flow features $X_{o} \in \mathbb{R}^{d \times T}$ at the snippet level, where $d$ denotes the feature dimension. 
We then concatenate them as $F=[ X_{r};X_{o}] \in \mathbb{R}^{2d \times T}$.
Each snippet is a short sequence of (\eg, 16 in our case) consecutive frames.}
While $F$ contains local spatio-temporal information, 
it lacks a global context critical for TAD.
We hence leverage the self-attention mechanism \cite{vaswani2017attention} to learn the global context. 
Formally, we set the $Q/K/V$ of a Transformer encoder
as the features $F/F/F$.
\textcolor{black}{
To model finer action details efficiently,
we consider multiple temporal scales in a hierarchy.
We start with the finest temporal resolution \textcolor{black}{(\eg, sampling $T=800$ snippets)},
which is progressively reduced via temporal pooling $P(\theta)$
with kernel size $k$, stride $s$ and padding $p$.
For efficiency, we first apply temporal pooling:
$\hat{Q}^{s} = P(Q;\theta_{Q}), 
\hat{K}^{s} = P(K;\theta_{K})$ and $\hat{V}^{s} = P(V;\theta_{V})$ with 
the scale $s \in \{1,2,4\}$.
The self-attention then follows as:
\begin{equation}\label{eqn_1}
A_{i}^{s} = F + softmax(\frac{F W_{\hat{Q}^{s}} ({F} W_{\hat{K}^{s}})^{\top}}{\sqrt{d}}) ({F} W_{\hat{V}^{s}}),
% % % % \vspace{-0.25cm}
\end{equation}
where $W_{\hat{Q}^{s}}, W_{\hat{K}^{s}}, W_{\hat{V}^{s}}$ are learnable parameters. In multi-head attention (MA) design, for each scale $s$ we combine a set of $n_{h}$ independent heads $A_{i}$ to form a richer learning process. The snippet embedding $E$ at scale $s$ is obtained as:
\begin{equation}\label{eqn_2}
E^{s} = MLP(\underset{MA}{\underbrace{[A_{1}^{s} \cdots A_{n_h}^{s}]}}) \ \in \ \mathbb{R}^{T^{s} \times C}.
% % % \vspace{-0.25cm}
\end{equation}
The {Multi-Layer Perceptron} (MLP) block has one fully-connected layer with residual skip connection. Layer norm is applied before both the MA and MLP block. 
We use $n_h=4$ heads by default.
}

\subsection{Parallel action classification and global segmentation masking}
\label{sec:model_head}
Our TAD head consists of two parallel branches:
one for multi-class action classification 
and 
the other for binary-class global segmentation mask inference.

% % \vspace{0.1cm}
\noindent\textbf{Multi-class action classification}
Given the $t$-th snippet $E^{s}(t) \in \mathbb{R}^{c}$ (i.e, the $t$-th column of $E^{s}$),
our classification branch predicts the probability $\bm{p}_t \in \mathbb{R}^{(K+1)\times 1}$ that
it belongs to one of $K$ target action classes or background.
This is realized by a 1-D convolution layer $H_c$ followed by 
a softmax normalization.
Since a video has been encoded into $T^{s}$ temporal snippets, the output of the classification branch can be expressed in column-wise as:
\begin{equation}\label{eqn_3}
\bm{P}^{s} : = softmax(H_c(E^{s})) \in \mathbb{R}^{(K+1) \times T^{s}}.
\end{equation}

\begin{figure}
% % % \vspace{-.15in}
\centering
    \includegraphics[scale=0.17]{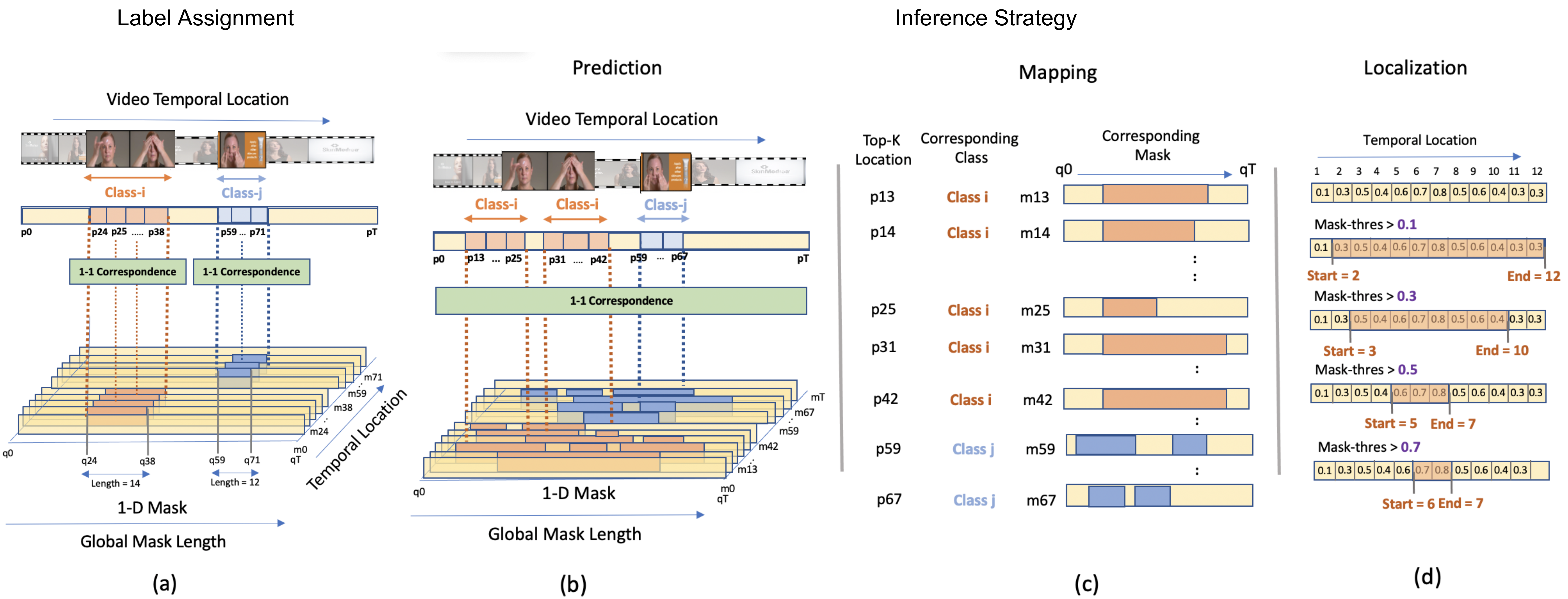}
    % \includegraphics[scale=0.46]{latex/img/ECCV22_GSM_fig3_v4.png}
    % % % \vspace{-.15in}
    \caption{Example of label assignment and model inference (see text for details).}
    \label{fig:label_infer}
% \end{minipage}
% % \vspace{-.15in}
\end{figure}

% % \vspace{0.1cm}
\noindent\textbf{Global segmentation mask inference} 
In parallel to the classification branch,
this branch aims to predict a global segmentation mask for each action instance of a video.
\textcolor{black}{
Each global mask is action instance specific and class agnostic.
For a training video, all temporal snippets of a single action instance are assigned with the same 1D global mask $\in \mathbb{R}^{T \times 1}$ for model optimization \textcolor{black} {(refer to Fig. \ref{fig:label_infer}(a)).}
}
For each snippet $E^{s}(t)$, 
it outputs a mask prediction $\bm{m}_t = [q_{1}, \cdots, q_{T}] \in \mathbb{R}^{T^{s} \times 1}$
with the $k$-th element $q_{k} \in [0, 1]$ indicating the foreground probability of $k$-th snippet conditioned on 
$t$-th snippet.
This process is implemented by a stack of three 1-D conv layers as:
\begin{equation}\label{eqn_4}
\bm{M}^{s} : = sigmoid(H_b(E^{s})) \in \mathbb{R}^{ T^{s} \times T^{s}}, 
\end{equation}
where the $t$-th column of $\bm{M}$ is the segmentation mask prediction at the $t$-th snippet.
With the proposed mask signal as learning supervision,
our {\shortmodelname} model can facilitate context-aware representation learning, which brings clear benefit on TAD accuracy (see Table \ref{tab:emb}).

% % \vspace{0.1cm}

\noindent\textcolor{black}{\noindent \textit{Remarks: } 
Actionness \cite{lin2019bmn,zhao2017temporal} is 
a popular localization method which predicts a single mask
in shape of $\in \mathbb{R}^{T \times 1}$.
There are several key differences between actionness and {\shortmodelname}: 
(1) Our per-snippet mask model
{\shortmodelname} focuses on a single action instance per snippet per mask
so that all the foreground parts of a mask
are intrinsically related; In contrast, actionness does not.
(2) {\shortmodelname} breaks the single multi-instance 1D actionness problem into a multiple 1D single-instance mask problem (refer to Fig. \ref{fig:label_infer}(a)).
This takes a divide-and-conquer strategy. 
By explicitly segmenting foreground instances at different temporal positions, {\shortmodelname} converts the regression based actionness problem into a position aware classification task. Each mask, associated with a specific time $t$, focuses on a single action instance. 
On the other hand, one action instance would be predicted
by multiple successive masks.
This predictive redundancy, simply removable by NMS,
provides rich opportunities for accurate detection.
(3) Whilst learning a 2D actionness map,
BMN \cite{lin2019bmn} relies on predicting 1D probability sequences which are highly noisy causing many false alarms. Further, its confidence evaluation cannot model the relations between candidates whilst our {\shortmodelname} can (Eq. \eqref{eq:pred_score}).
Lastly, our experiments in Table \ref{tab:maskd} validate the superiority of {\shortmodelname} over actionness learning.
}

% % \vspace{-0.1in}
\subsection{Model Training}
\noindent{\bf Ground-truth labels. }
\textcolor{black}{To train {\shortmodelname}, the ground-truth needs to be arranged into the designed format. Concretely, given a training video with temporal intervals and class labels (Fig.~\ref{fig:label_infer}(a)), we label all the snippets (orange or blue squares) of a single action instance with the same action class. All the snippets off from action intervals
are labeled as background. For an action snippet of a particular instance, its global mask is defined as the video-length
binary mask of that action instance. 
Each mask is action instance specific.
All snippets of a specific action instance share
the same mask. For instance, all orange snippets (Fig. \ref{fig:label_infer}(a)) are assigned with a $T$-length mask (eg. $m_{24}$ to $m_{38}$) with one in the interval of $[q24,q38]$.
}

% % \vspace{0.1cm} 
\noindent\textbf{Learning objectives. } 
The classification branch is trained by a combination of a cross-entropy based focal loss %$L_{focal}$ 
and a class-balanced logistic regression loss %$L_{lr}$
\cite{dong2019single}. 
For a training snippet, we denote $y$ the ground-truth class label, $\bm{p}$ the classification output,
and \textcolor{black}{$\bm{r}$} the per-class regression output \textcolor{black}{obtained by applying sigmoid on top of $H_{c}$ in Eq. \eqref{eqn_3}} (discarded at inference time).
% ${\mathbb{O}}_{c}$, 
The loss of the classification branch is then written as:
\begin{equation}\label{eqn6}
 L_{c} = \lambda_{1}(1-\bm{p}(y))^{\gamma }\log(\bm{p}_y) + (1-\lambda_{1})
\big( \log(\bm{r}_y) 
- \frac{\alpha}{|\mathcal{N}|} \sum_{k\epsilon \mathcal{N}}(\log( 1-\bm{r}(k))) 
\big),   
\end{equation}
where $\gamma=2$ is a focal degree parameter,
$\alpha=10$ is a class-balancing weight,
and $\mathcal{N}$ specifies a set of hard negative classes
at size of $K/10$ where $K$ is the toTAD action class number.
We set the loss trade-off parameter $\lambda_{1}=0.4$. 

\textcolor{black}{For training the segmentation mask branch, we combine a novel boundary IOU (bIOU) loss and the dice loss in \cite{milletari2016v} to model two types of structured consistency respectively: mask boundary consistency and inter-mask consistency. Inspired by the boundary IOU metric \cite{cheng2021boundary},
bIOU is designed particularly to penalize incorrect temporal boundary prediction w.r.t. the ground-truth segmentation mask.
Formally, for a snippet location, 
we denote $\bm{m} \in \mathbb{R}^{T\times 1}$ the predicted segmentation mask, and
$\bm{g} \in \mathbb{R}^{T\times 1}$ the ground-truth mask.
The overall segmentation mask loss is formulated as:
} 
\begin{equation}
\centering \small
\begin{aligned}
    L_m = 1 - \Big(\frac{ \cap(m,g)}{\cup(m,g)} + \frac{1}{\cap(m,g) + \epsilon} \frac{\left \| \bm{m} - \bm{g} \right \|_{2}}{c}\Big)
      +  
    \lambda_2
    \Big( 1 - 
    \frac{\bm{m}^\top \bm{g}}
    {\sum_{t=1}^{T} \big(\bm{m}(t)^2 + \bm{g}(t)^2 \big)} 
    \Big),
\end{aligned}\label{eq:mask_loss}
\end{equation}
where $\cap(m,g) = \Phi(\bm{m}) \cap \Phi(\bm{g})$ and $\cup(m,g) = \Phi(\bm{m}) \cup \Phi(\bm{g})$, $\Phi(\cdot)$ represents a kernel of size $k$ ($7$ in our default setting, see more analysis in Suppl.) used as a differentiable morphological erosion operation \cite{riba2020kornia} on a mask and $c$ specifies the ground-truth mask length.
In case of no boundary overlap between the predicted and ground-truth masks, 
% we penalize more using
we use the normalized $L_2$ loss.
The constant $\epsilon=e^{-8}$ is introduced for numerical stability.
We set the weight $\lambda_2 = 0.4$.

 \begin{figure}[t]
% % \vspace{-.15in}
\centering
\includegraphics[scale=0.44]{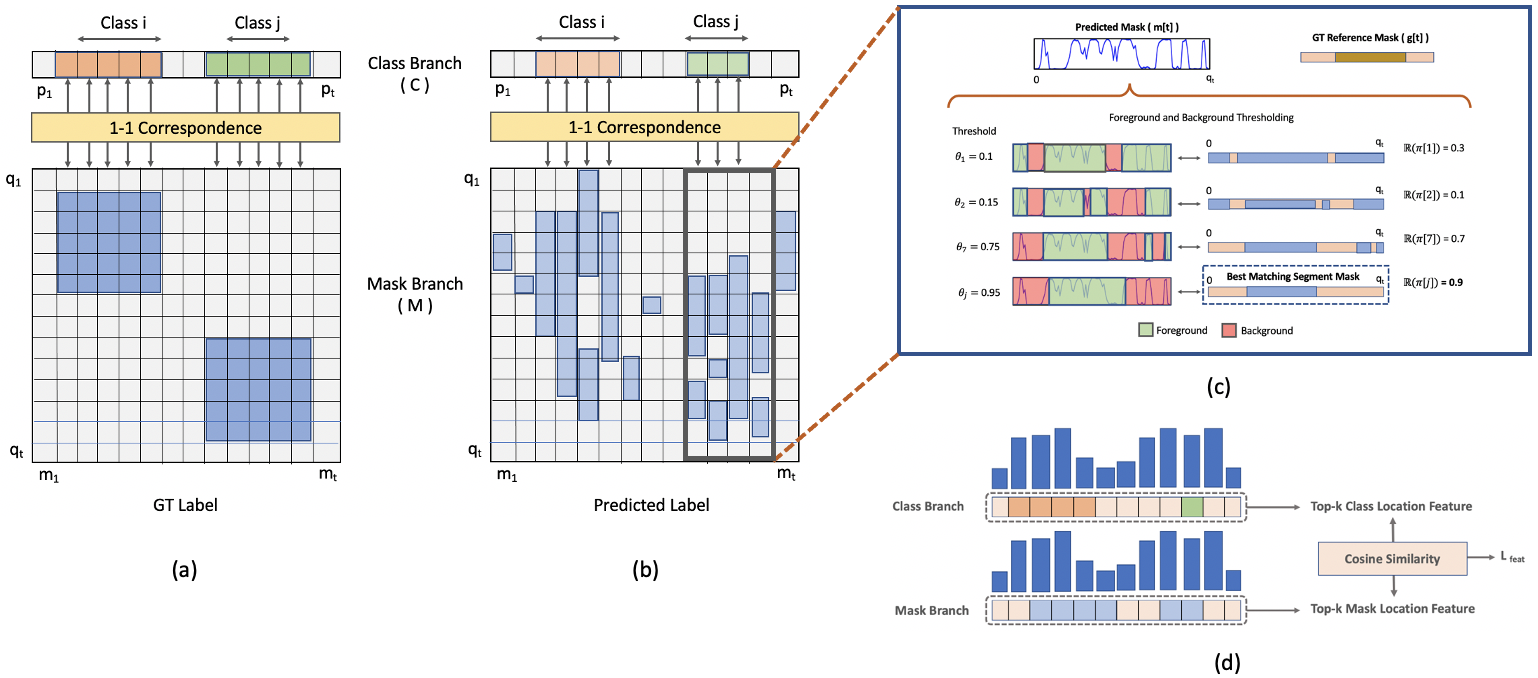}
% % % \vspace{-.15in}
\caption{\color{black}
An example of (a) ground-truth labels and (b) prediction along with an illustration of exploring (c) mask predictive redundancy (Eq. \eqref{eq:pred_score}) (d) classification-mask consistency ((Eq. \eqref{eq:clmask}).
}
\label{fig:toy_ex}
% \end{minipage}
% % \vspace{-.15in}
\end{figure}

\noindent{\em Mask predictive redundancy }
\textcolor{black}{Although the mask loss Eq. \eqref{eq:mask_loss} above treats the global mask as a 2D binary mask prediction problem, 
it cannot always regulate the behaviour of individual 1D mask within an action instance.}
\textcolor{black}{
Specifically, for a predicted mask $m_t$ at time $t$,
thresholding it at a specific threshold $\theta_j \in {\Theta}$
can result in {binarized segments of foreground and background}:
$\pi[j] = \{(q_{s}^{i},q_{e}^{i},z_{i})\}_{i=1}^{L}$ where $q^{i}_{s}$ and $q^{i}_{e}$ denotes the start and end of $i$-th segment, and $z_{i} \in \{0,1\}$ indicates background or foreground. 
{For a mask corresponding to an action snippet, ideally  at least one of these $\{\pi[j]\}$ should be closer to the ground truth.}
To explore this redundancy, we define a prediction scoring criterion
with the outer-inner-contrast \cite{shou2018autoloc,lee2021learning} as follows:
}
\begin{equation}
\begin{aligned}\footnotesize
  \mathbb{R}(\pi[j]) =  \frac{1}{L}\sum_{i=1}^{L} \left (    \underbrace{\frac{1}{l_i}\sum_{r=q^{i}_s}^{q^{i}_e}u_i(r) }_{\text{inside}}
- \underbrace{\frac{1}{\left \lceil \delta l_i  \right \rceil + \left \lceil \delta l_i  \right \rceil }\left ( \sum_{r=q^{i}_s - \left \lceil \delta l_i \right \rceil}^{q^{i}_s - 1} u_i(r) + \sum_{r=q^{i}_e +1}^{q^{i}_e + \left \lceil \delta l_i \right \rceil} u_i(r) \right )}_{_{\text{outside}} } \right )
 \\
  where \hspace{0.2in}u_i(r) = \left\{\begin{matrix}
m_{t}[r], &\text{if}\;\; z_i=1 \; \text{(\ie, foreground)}\\ 
 1 - m_{t}[r], & \text{otherwise}
\end{matrix}\right.
 \end{aligned}
 \label{eq:pred_score}
\end{equation}
\textcolor{black}{$l_i = q^{i}_e - q^{i}_s + 1$ is the temporal length of $i$-th segment, $\delta$ is a weight hyper-parameter which is set to 0.25.}
\textcolor{black}{
We obtain the best prediction with the maximal score as 
$j^{*} = argmax(\mathbb{R}(\pi[j]))$ (see Fig. \ref{fig:toy_ex}(c)).
Higher $\mathbb{R}(\pi[j^{*}])$ means 
a better prediction.
To encourage this best prediction,
we design a prediction promotion loss function:
\begin{equation}
    L_{pp} =  \big(1 - \mathbb{R}(\pi[j^{*}])\big)^{\beta} \left \| m_{t} - g_{t} \right \|_{2},
\end{equation}
where we set $\beta=2$ for penalizing lower-quality prediction stronger.
We average this loss across all snippets of each action instance per training video.
}

{\color{black}
\noindent{\em Classification-mask consistency }
In {\shortmodelname}, there is structural consistency {in terms of foreground} between class and mask labels by design (Fig. \ref{fig:toy_ex}(a)).
To leverage this consistency, we formulate a feature consistency loss as:
\begin{equation}
     L_{fc} = 1 - \texttt{cosine}\Big( 
    \hat{F}_{clf},
    \hat{F}_{mask}
    \Big),
\label{eq:clmask}
\end{equation}
where $\hat{F}_{clf} = topk(argmax((P_{bin}*E_{p})[:K,:]))$ is the features obtained from the top scoring foreground snippets obtained from the thresholded classification output $P_{bin} := \eta(P - \theta_{c})$
with $\theta_{c}$ the threshold and $E_{p}$ obtained by passing the embedding $E$ into a 1D conv layer for matching the dimension of $P$.
The top scoring features from the mask output $M$ are obtained similarly as:
$\hat{F}_{mask} = topk(\sigma(1DPool(E_{m}*M_{bin})))$
where $M_{bin} := \eta(M - \theta_{m})$ is a binarization of mask-prediction $M$, $E_{m}$ is obtained by passing the embedding $E$ into a 1D conv layer for matching the dimension of $M$, $*$ is element-wise multiplication, $\eta(.)$ is the binarization function, and $\sigma$ is sigmoid activation.
Our intuition is that the foreground features should be closer and consistent after the classification and masking process \textcolor{black}{(refer to Fig. \ref{fig:toy_ex}(d))}.
}

\noindent\textit{Overall objective } 
The overall objective loss function for training {\shortmodelname} is defined as:
$L = L_{c} + L_{m} + L_{pp} + L_{fc}$. This loss is calculated for each temporal scale $s$ and finally aggregated over all the scales.

\subsection{Model Inference}
Our model inference is similar as existing TAD methods \cite{lin2019bmn,xu2020g}.
\textcolor{black}{Given a test video, at each temporal scale $s$ the action instance predictions are first generated separately based on the classification $\bm{P}^{s}$ and mask $\bm{M}^{s}$ predictions and combined for the following post-processing.
Starting with the top %$M_1$ 
scoring snippets from $\bm{P}$ (Fig \ref{fig:label_infer}(b)),
we obtain their segmentation mask predictions (Fig \ref{fig:label_infer}(c))
by thresholding the corresponding columns of $\bm{M}$ (Fig \ref{fig:label_infer}(d)).
To generate sufficient candidates,
we apply multiple thresholds ${\Theta}=\{\theta_i\}$
to yield action candidates with varying lengths and confidences.
% and combine all the outputs.
%
For each candidate, we compute its confidence score $sc_{final}$
by multiplying the classification score \textcolor{black}{(obtained from the corresponding top-scoring snippet in $P$)}
and the \textcolor{black}{segmentation mask score (\ie,  the {\em mean} predicted foreground segment in $M$)}.
Finally, we apply SoftNMS \cite{bodla2017soft} on top scoring candidates to obtain the final predictions.}

\section{Experiments}
\label{sec:exp}
\noindent{\bf Datasets }
We conduct extensive experiments on two popular TAD benchmarks.
(1) \textit{ActivityNet-v1.3}~\cite{caba2015activitynet} has 19,994 videos from 200 action classes. We follow the standard setting %~\cite{lin2018bsn} 
to split all videos into training, validation and testing subsets 
in ratio of 2:1:1.
% THUMOS14~\cite{jiang2014thumos} and ActivityNet1.3~\cite{caba2015activitynet}.
(2) \textit{THUMOS14}~\cite{idrees2017thumos} has 200 validation videos and 213 testing videos from 20 categories with labeled temporal boundary and action class.

% % \vspace{0.1cm}
\noindent{\bf Implementation details } 
We use two pre-extracted encoders for feature extraction, for fair comparisons with previous methods. 
% For ActivityNet-v1.3, % One we adopt the
One is a fine-tuned two-stream model \cite{lin2019bmn}, with downsampling ratio 16 and stride 2. Each video's feature sequence $F$ is rescaled to $T = 800/1024$ snippets for AcitivtyNet/THUMOS using linear interpolation. %For THUMOS14, 
The other is %are extracted using 
Kinetics pre-trained I3D model \cite{carreira2017quo} with a downsampling ratio of 5. 
Our model is trained for 15 epochs using Adam with learning rate of {$10^{-4}/10^{-5}$ for AcitivityNet/THUMOS respectively. The batch size is set to 50 for ActivityNet and 25 for THUMOS}. 
\textcolor{black}{For classification-mask consistency, the threshold $\theta_{m}/\theta_{p}$ is set to $0.5/0.3$ and in top$-k$ to 40}. In testing, 
we set the threshold set for mask $\Theta=\{0.1\sim 0.9\}$ with step $0.05$. We use the same set of threshold $\Theta$ for mask predictive redundancy during training.

% % \vspace{-0.20in}
\begin{table*}[t]
% % \vspace{-0.20in}
\footnotesize
\centering
\renewcommand{\tabcolsep}{2pt}
\renewcommand{\arraystretch}{0.8}
% \setlength{\tabcolsep}{0.25cm}
% \resizebox{2\columnwidth}{!}
{
\resizebox{\textwidth}{!}{%
\begin{tabular}{@{}c|l|c|cccccc|cccc@{}}
\toprule

% \toprule
\multirow{2}{*}{\textbf{Type}} &
  \multirow{2}{*}{\textbf{Model}} &
  \multirow{2}{*}{\textbf{Bkb}} &
  \multicolumn{6}{c|}{\textbf{THUMOS14}} &
  \multicolumn{4}{c}{\textbf{ActivityNet-v1.3}} \\ \cmidrule(l){4-13} 
                              &             &     & 0.3  & 0.4  & 0.5  & 0.6  & 0.7  & Avg. & 0.5           & 0.75          & 0.95 & Avg.          \\ \midrule 
                              \multirow{8}{*}{Anchor} & R-C3D        & C3D & 44.8 & 35.6 & 28.9 & -    & -    & -    & 26.8          & -             & -    & -             \\ \cmidrule(l){4-13} 
                              & TAD        & I3D & 53.2 & 48.5 & 42.8 & 33.8 & 20.8 & 39.8 & 38.2          & 18.3          & 1.3  & 20.2          \\ \cmidrule(l){4-13} 
                              & GTAN        & P3D & 57.8 & 47.2 & 38.8 & -    & -    & -    & 52.6          & 34.1          & 8.9  & 34.3          \\ \cmidrule(l){4-13}
                              & PBR-Net       & I3D & 58.5 & 54.6 & 51.3 & 41.8    & 29.5    & -    & 53.9          & 34.9           & 8.9  & 35.0          \\ \cmidrule(l){4-13}
                              & MUSES        & I3D & \bf 68.9 & \bf 64.0 & 56.9 & 46.3 &  31.0 & \bf 53.4 & 50.0          & 34.9          & 6.5   & 34.0         \\ \cmidrule(l){4-13}
                               & VSGN     & I3D  & 66.7 & 60.4 & 52.4 & 41.0 & 30.4 & 50.1 & 52.3           & 36.0          & 8.3  & 35.0          \\ 
                              \midrule
                              \multirow{16}{*}{Actn}  & BMN        & TS  & 56.0 & 47.4 & 38.8 & 29.7 & 20.5 & 38.5 & 50.1          & 34.8          & 8.3  & 33.9          \\ \cmidrule(l){4-13} 
                              & DBG        & TS  & 57.8 & 49.4 & 42.8 & 33.8 & 21.7 & 41.1 & -             & -             & -    & -             \\ \cmidrule(l){4-13} 
                              & G-TAD      & TS  & 54.5 & 47.6 & 40.2 & 30.8 & 23.4 & 39.3 & 50.4          & 34.6          & 9.0  & 34.1          \\ \cmidrule(l){4-13} 
                              & BU-TAL      & I3D & 53.9 & 50.7 & 45.4 & 38.0 & {28.5} & 43.3 & 43.5          & 33.9          & 9.2  & 30.1          \\ \cmidrule(l){4-13} 
                              & BSN++      & TS  & 59.9 & 49.5  & 41.3  & 31.9  & 22.8 & - &  51.2           & 35.7          & 8.3   & 34.8          \\ \cmidrule(l){4-13}
                              & GTAD+CSA      & TS  & 58.4  & 52.8 & 44.0 & 33.6 & 24.2 & 42.6 & 51.8           & \bf 36.8          & 8.7  & 35.7          \\ \cmidrule(l){4-13}
                    
                              & BC-GNN      & TS  & 57.1 & 49.1 & 40.4 & 31.2 & 23.1 & 40.2 & 50.6          & 34.8          & 9.4  & 34.3          \\
                              \cmidrule(l){4-13} 
                              & TCANet      & TS  & 60.6 & 53.2 & 44.6 & 36.8 & 26.7 & -    & 52.2          & 36.7 & 6.8  & 35.5          \\
                              \cmidrule(l){4-13}
                              & ContextLoc     & I3D  & 68.3 & 63.8 & 54.3 & 41.8 & 26.2 & -    & 56.0         & 35.2  & 3.5  & 34.2          \\
                              \cmidrule(l){4-13}
                              \cmidrule(l){4-13}
                              & RTD-Net         & I3D & 68.3 & 62.3 & 51.9 & 38.8 & 23.7 & -   & 47.2          & 30.7          & 8.6  & 30.8          \\  
                              
                              \midrule
\multirow{2}{*}{Mixed}        & A2Net      & I3D & 58.6 & 54.1 & 45.5 & 32.5 & 17.2 & 41.6 & 43.6          & 28.7          & 3.7  & 27.8          \\ \cmidrule(l){4-13} 
                              & GTAD+PGCN  & I3D & 66.4 & 60.4 & 51.6 & 37.6 & 22.9 & 47.8 & -             & -             & -    & -             \\ \midrule
% \multirow{4}{*}
\multirow{3}{*}{ PF}  
                            
                            &\bf {\shortmodelname} (Ours) & I3D & 68.6   & 63.8   & \bf 57.0   & \bf 46.3   & \textbf{31.8}   & 52.8  
                              & \textbf{56.3} & \bf 36.8 &\bf \textbf{9.6}  & \textbf{36.5} \\
                            
                              & \bf {\shortmodelname} (Ours)       & TS  & \textbf{61.4}   & \bf 52.9   & \textbf{46.5}   & \textbf{38.1}   & \bf 27.0   & \textbf{44.0}
                              & \textbf{53.7}            & 36.1            & \bf 9.5   & \textbf{35.9}\\ 
                              \bottomrule
\end{tabular}
}
\caption{Performance comparison with state-of-the-art methods on THUMOS14 and ActivityNet-v1.3.
The results are measured by mAP at different IoU thresholds, and average mAP in {[}0.3 : 0.1 : 0.7{]} on THUMOS14 and {[}0.5 : 0.05 : 0.95{]} on ActivityNet-v1.3.
Actn = Actioness;
PF = Proposal Free;
Bkb = Backbone.
% $\dagger$ is refined using UntrimmedNet Classifier.
}
\label{tab:tab_1}
% \vspace{-0.30in}
}
\end{table*}

\subsection{Main Results}

\noindent\textbf{Results on ActivityNet } 
From Table \ref{tab:tab_1},
we can make the following  observations:
(1) {\shortmodelname} with I3D feature achieves the best result in average mAP. Despite the fact that our model is much simpler in architecture design compared to the existing alternatives. This validates our assumption that with proper global context modeling, explicit proposal generation is not only redundant but also less effective. 
(2) When using the relatively weaker two-stream (TS) features,
our model remains competitive and even surpasses I3D based
BU-TAL \cite{zhao2020bottom}, A2Net \cite{yang2020revisiting} and the very recent  ContextLoc \cite{zhu2021enriching} and MUSES \cite{liu2021multi} by a significant margin.
\textcolor{black}{{\shortmodelname} also surpasses a proposal refinement and strong G-TAD based approach CSA \cite{sridhar2021class} 
on avg mAP.}
\textcolor{black}{(3) Compared to RTD-Net which employs an architecture similar to object detection Transformers, our {\shortmodelname} is significantly superior. This validates our model formulation in exploiting the Transformer for TAD.}

% % \vspace{0.1cm}
\noindent \textbf{Results on THUMOS14 } 
Similar conclusions can be drawn in general on THUMOS from Table \ref{tab:tab_1}. 
\textcolor{black}{When using TS features, {\shortmodelname} achieves again the best results, beating strong competitors like TCANet \cite{wang2021temporal}, CSA \cite{sridhar2021class} by a clear margin. % of xx\%.
There are some noticeable differences:
(1) We find that I3D is now much more effective than two-stream (TS),
\eg, 8.8\% gain in average mAP over TS with {\shortmodelname}, compared with 0.6\% on ActivityNet. This is mostly likely caused by the distinctive characteristics of the two datasets in terms of action instance duration and video length.
(2) Our method achieves the second best result with marginal edge behind MUSES \cite{liu2021multi}.
% in only 0.6 and 0.7 mAP. 
This is partly due to that MUSES benefits from additionally tackling the scene-changes. \textcolor{black}{(3) Our model achieves the best results in stricter IOU metrics (\eg, IOU@0.5/0.6/0.7) consistently using both TS and I3D features, verifying the effectiveness of solving mask redundancy.}
}

% % \vspace{0.1cm}
\noindent{\bf Computational cost comparison }
One of the key motivations to design a proposal-free TAD model is to reduce the model training and inference cost.
For comparative evaluation, we evaluate
% wraptable commented out 
{\shortmodelname} against two representative and recent TAD methods (BMN \cite{lin2019bmn} and G-TAD \cite{xu2020g})
using their released codes.
All the methods are tested on the same machine with one Nvidia 2080 Ti GPU.
We measure the convergence time in training
and average inference time per video in testing.
{The two-stream video features are used.}
It can be seen in Table \ref{tab:speed} that our {\shortmodelname}
is drastically faster, \eg, $20/25\times$ for training and clearly quicker -- $1.6/1.8\times$ for testing in comparison to G-TAD/BMN, respectively.
We also notice that our {\shortmodelname} needs less epochs to converge.
Table \ref{tab:param} also shows that our {\shortmodelname} has the smallest FLOPs and the least parameter number.

\begin{table}[t]
    % \caption{Global caption}
    \begin{minipage}{.45\linewidth}
    \setlength{\tabcolsep}{5pt}
      \caption{Analysis of model training and test cost.}
      \centering
      \label{tab:speed}
        \begin{tabular}{c|c|c|c}
        % \toprule
        \toprule
        Model & Epoch & Train & Test
        \\ \midrule
        
        BMN %\cite{lin2019bmn}       
        % & 3402
        & 13
        & 6.45 hr       & 0.21 sec            \\
        G-TAD %\cite{xu2020g}         %& 1410   
        & 11
        & 4.91 hr     & 0.19 sec         \\ 
        \midrule
        \textbf{{\shortmodelname}} %& \textbf{9.5} 
        & \bf 9
        & \textbf{0.26} hr & \textbf{0.12} sec \\ \bottomrule
        \end{tabular}
    \end{minipage}%
    \hfill
    \begin{minipage}{.5\linewidth}
      \centering
      \setlength{\tabcolsep}{6pt}
        \caption{Analysis of model parameters \# and FLOPs.}
        \begin{tabular}{ll}
          \begin{tabular}{@{}c|c|c@{}}
            % \toprule
            \toprule
            Model & Params (in M) & FLOPs (in G)
                                                             \\ \midrule
            BMN                    & 5.0                            & 91.2                          \\
            GTAD                   & 9.5                            & 97.2                          \\ \hline
            \textbf{{\shortmodelname}}    & \textbf{6.2}                   & \textbf{17.8}                  \\ \bottomrule
            \end{tabular}
        \end{tabular}
        \label{tab:param}
    \end{minipage} 
\end{table}

\subsection{Ablation study and further analysis}
\label{sec:ablation}
\noindent{\bf Transformers vs. CNNs }
We compare our multi-scale Transformer with CNN for snippet embedding.
We consider two CNN designs:
(1) a 1D CNN
with 3 dilation rates 
(1, 3, 5) each with 2 layers,
 and 
(2) a multi-scale MS-TCN \cite{farha2019ms}, 
and (3) a standard single-scale Transformer \cite{vaswani2017attention}.
Table \ref{tab:emb} shows that 
the Transformers are clearly superior to both 1D-CNN and a relatively stronger MS-TCN. This suggests that our global segmentation mask learning is more compatible with self-attention models due to stronger contextual learning capability.
% \begin{table}[h]
% \setlength{\tabcolsep}{12pt}
%     \caption{Ablation of Transformer vs. CNN on ActivityNet.}
%       \centering
%       \label{tab:emb}
%         \begin{tabular}{l|c|c}
% \toprule
% % \toprule
% \multirow{2}{*}{\textbf{Embedding Network}} &
%   \multicolumn{2}{c}{\textbf{mAP}} \\ 
%   \cmidrule(l){2-3}
% %   \midrule
% & \textbf{0.5}  & \textbf{Avg}  \\ \midrule
% 1D CNN  & 46.8          & 26.4          \\
% MS-TCN  & 53.1          & 33.8          \\
% Transformer & 55.8 & 36.1 \\
% \textbf{MS-Transformer (Ours)} & \textbf{56.3} & \textbf{36.5} \\
% \bottomrule
% \end{tabular}
% \end{table}
\noindent \textcolor{black}{Besides, multi-scale learning with Transformer gives $0.4\%$ gain in avg mAP validating the importance of larger snippets. As shown in Table \ref{tab:pool}, the gain almost saturates from 200 snippets,
and finer scale only increases the computational cost.}
% \begin{table}[]
% \setlength{\tabcolsep}{10pt}
%   \centering
%         \caption{Ablation on snippet embedding design and multiple temporal scales.}
%       \begin{tabular}{@{}c|c|c|c|cc@{}}
% \toprule
% \multirow{2}{*}{Scale} &
%   \multirow{2}{*}{Snippet} &
%   \multirow{2}{*}{\begin{tabular}[c]{@{}c@{}}Params\\ (in M)\end{tabular}} &
%   \multirow{2}{*}{\begin{tabular}[c]{@{}c@{}}Inference\\ (in sec)\end{tabular}} &
%   \multicolumn{2}{c}{mAP} \\ \cmidrule(l){5-6} 
%   &             &     &      & \multicolumn{1}{c|}{0.5}  & Avg  \\ \midrule
% \{1\} & 100         & 2.9 & 0.09 & \multicolumn{1}{c|}{55.8} & 36.1 \\ \midrule
% \textbf{\{1,2\}} & \textbf{100,200}     & \textbf{6.2} & \textbf{0.12} & \multicolumn{1}{c|}{56.3} & \textbf{36.5} \\
% \{1,2,4\} & 100,200,400 & 9.8 & 0.16 & \multicolumn{1}{c|}{\textbf{56.5}} & 36.4 \\ \bottomrule
% \end{tabular}
%         \label{tab:pool} 
% \end{table}

\begin{figure}[h]
    \centering
    \includegraphics[scale=0.14]{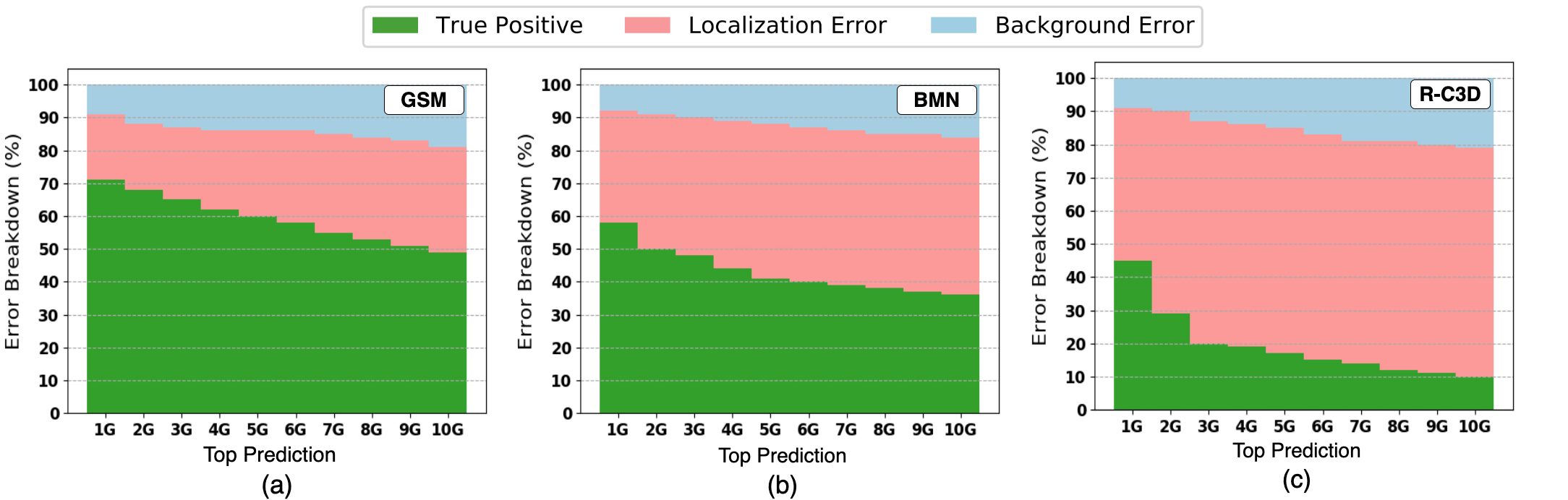}
    % % \vspace{-0.1in}
    \caption{False positive profile of {\shortmodelname}, BMN and R-C3D on ActivityNet.
  We use top up-to 10$G$ predictions per video, where $G$ is the number of ground truth action instances.}
    \label{fig:error_prof}
    % % \vspace{-0.1in}
\end{figure}

\begin{table}[!htb]
    % \caption{Global caption}
    \begin{minipage}{.45\linewidth}
       \caption{Ablation of Transformer  \\vs. CNN on ActivityNet.}
      \centering
      \label{tab:emb}
        \begin{tabular}{l|c|c}
\toprule
% \toprule
\multirow{2}{*}{\textbf{Network}} &
  \multicolumn{2}{c}{\textbf{mAP}} \\ 
  \cmidrule(l){2-3}
%   \midrule
& \textbf{0.5}  & \textbf{Avg}  \\ \midrule
1D CNN  & 46.8          & 26.4          \\
MS-TCN  & 53.1          & 33.8          \\
Transformer & 55.8 & 36.1 \\
\textbf{MS-Transformer} & \textbf{56.3} & \textbf{36.5} \\
\bottomrule
\end{tabular}
    \end{minipage}%
    \hfill
    \begin{minipage}{.55\linewidth}
      \centering
       \caption{Ablation on snippet embedding design and multiple temporal scales.}
       \begin{tabular}{@{}c|c|c|c|cc@{}}
\toprule
\multirow{2}{*}{Scale} &
  \multirow{2}{*}{Snippet} &
  \multirow{2}{*}{\begin{tabular}[c]{@{}c@{}}Params\\ (in M)\end{tabular}} &
  \multirow{2}{*}{\begin{tabular}[c]{@{}c@{}}Infer\\ (in sec)\end{tabular}} &
  \multicolumn{2}{c}{mAP} \\ \cmidrule(l){5-6} 
  &             &     &      & \multicolumn{1}{c|}{0.5}  & Avg  \\ \midrule
\{1\} & 100         & 2.9 & 0.09 & \multicolumn{1}{c|}{55.8} & 36.1 \\ \midrule
\textbf{\{1,2\}} & \textbf{100,200}     & \textbf{6.2} & \textbf{0.12} & \multicolumn{1}{c|}{56.3} & \textbf{36.5} \\
\{1,2,4\} & 100,200,400 & 9.8 & 0.16 & \multicolumn{1}{c|}{\textbf{56.5}} & 36.4 \\ \bottomrule
\end{tabular}
        \label{tab:pool} 
    \end{minipage} 
\end{table}
\noindent{\bf Proposal-based vs. proposal-free}
We compare our proposal-free {\shortmodelname} with conventional proposal-based TAD methods BMN \cite{bai2020boundary} \textcolor{black}{(anchor-free)} and R-C3D \cite{xu2017r} \textcolor{black}{(anchor-based)} via false positive analysis \cite{alwassel2018diagnosing}. We sort the predictions by the scores and take the top-scoring predictions per video. Two major errors of TAD are considered:
\textcolor{black}{(1) {\em Localization error}, which is defined as when a proposal/mask is predicted as foreground, has a minimum tIoU of 0.1 but does not meet the tIoU threshold. 
(2) {\em Background error}, which happens when a proposal/mask
is predicted as foreground but its tIoU with ground truth instance is smaller than 0.1.
In this test, we use ActivityNet. 
\textcolor{black}{We observe in Fig. \ref{fig:error_prof} that {\shortmodelname} has the most true positive samples at every
amount of predictions. The proportion of localization error
with {\shortmodelname} is also notably smaller,
which is the most critical metric for improving average mAP \cite{alwassel2018diagnosing}.
This explains the gain of {\shortmodelname} over BMN and R-C3D.
}
}
\noindent{\bf Direction of improvement analysis }
Two subtasks are involved in TAD -- temporal localization and action classification, each of which would affect the final performance. 
Given the two-branch design in {\shortmodelname},
the performance effect of one subtask can be individually examined
by simply assigning ground-truth to the other subtask's output at test time.
From Table \ref{tab:imp_anal}, the following observations can be made:
(1) There is still a big scope for improvement  on both subtasks.
(2) Regarding the benefit from the improvement from the other subtask, the classification subtask seems to have the most to gain at mAP@0.5, 
whilst the localization task  can benefit more on the average mAP metric.
Overall, this analysis suggests that further improving the efficacy on the classification subtask would be more influential to the final model performance. 
% \vspace{-0.20in}
% \begin{table}[h]
%   \centering
%   \setlength{\tabcolsep}{12pt}
%         \caption{Improvement analysis of {\shortmodelname} on ActivityNet.}
%         \begin{tabular}{l|c|c}
%         % \toprule
%         \toprule
%         \multirow{2}{*}{\textbf{Model}} & \multicolumn{2}{c}{\textbf{mAP}} \\ \cmidrule(l){2-3} 
%         & \textbf{0.5}    & \textbf{Avg}    \\ \midrule
%         {\shortmodelname} (full) &     56.3  &     36.5           \\ \midrule
%         + Ground-truth class &       61.0          &        43.8 ( $\uparrow 7.3\%$)         \\
%         + Ground-truth mask   &        69.2         &          48.5 ( $\uparrow 12.0\%$)      \\ \bottomrule
%         \end{tabular}
%         \label{tab:imp_anal}
% \end{table}
% \vspace{-0.10in}

\noindent{\bf Analysis of components }
We can see in Table \ref{tab:designchoice} that without the proposed segmentation mask branch, the model will degrade significantly, \eg, a drop of 7.6\% in average mAP. This is due to its fundamental capability of modeling the global temporal structure of action instances and hence yielding better action temporal intervals. 
Further, for {\shortmodelname} we use a pre-trained UntrimmedNet (UNet) \cite{wang2017untrimmednets} as an external classifier instead of using the classification branch, resulted in a 2-stage method.
This causes a performance drop of 4.7\%, suggesting
\textcolor{black}{that both classification and mask branches 
are critical for model accuracy and efficiency.}

% \vspace{-0.10in}
\begin{table}[h]
	\begin{minipage}{0.45\linewidth}
		\caption{Analysis of {\shortmodelname}'s two branches on ActivityNet.}
      \centering
      \label{tab:designchoice}
        \begin{tabular}{@{}c|c|c@{}}
        % \toprule
        \toprule
        \multirow{2}{*}{\textbf{Model}}  & \multicolumn{2}{c}{\textbf{mAP}} \\ \cmidrule(l){2-3} 
                                & \textbf{0.5}         & \textbf{Avg}        \\ \midrule
        {\shortmodelname}(Full)               &      56.3       &     36.5       \\ \midrule
        w/o Mask Branch         &      45.8       &    28.9        \\
        w/o Class Branch + UNet &     49.7        &     31.8       \\ \bottomrule
        \end{tabular}
	\end{minipage}\hfill
	\begin{minipage}{0.50\linewidth}
% 		\captionof{figure}{Error analysis on THUMOS.}
% 		\includegraphics[height=1.2in,width=\columnwidth]{latex/img/ECCV22_GSMT_fig7_v2.png}
% 		\label{fig:segerr}
% % 		% \vspace{-0.20in}
\caption{Improvement analysis of {\shortmodelname} on ActivityNet.}
        \begin{tabular}{l|c|c}
        % \toprule
        \toprule
        \multirow{2}{*}{\textbf{Model}} & \multicolumn{2}{c}{\textbf{mAP}} \\ \cmidrule(l){2-3} 
        & \textbf{0.5}    & \textbf{Avg}    \\ \midrule
        {\shortmodelname} (full) &     56.3  &     36.5           \\ \midrule
        + Ground-truth class &       61.0          &        43.8 ( $\uparrow 7.3\%$)         \\
        + Ground-truth mask   &        69.2         &          48.5 ( $\uparrow 12.0\%$)      \\ \bottomrule
        \end{tabular}
        \label{tab:imp_anal}
	\end{minipage}
\end{table}
% % % \vspace{0.1cm}
% \noindent{\bf Effects of action instance duration }
% {We additionally evaluate how the model performance is affected by the duration of action instances on THUMOS.
% We compare our proposal-free method with a proposal based approach BMN
% \cite{lin2019bmn}. 
% % Fig. \ref{fig:segerr} shows the trend of 
% We measure segmentation error between the ground-truth and temporal prediction both with $L_1$ norm) against the ground-truth normalized duration. As seen in Fig. \ref{fig:segerr}, our {\shortmodelname} yields lower segmentation error than BMN particularly for shorter action instances w.r.t. the whole video length. 
% }

\noindent\textbf{Global mask design}
\textcolor{black}{
We compare our global mask with previous 1D actionness mask  \cite{lin2019bmn,xu2020boundary}.
We integrate actionness with {\shortmodelname} by reformulating the mask branch to output 1D actionness. %head of mask branch. 
From the results in Table~\ref{tab:maskd}, we observe a significant performance drop of $11.5\%$ in mAP@0.5 IOU. One reason is that the number of action candidates generated by actionness is drastically limited,
leading to poor recall. Additionally, we
visualize the cosine similarity scores of all snippet feature pairs on a random ActivityNet val video. As shown in Fig.~\ref{fig:cos}, 
our single-instance mask (global mask) design learns more discriminating feature representation with larger separation between background and action, as compared to multi-instance actionness design.
This validates the efficacy of our design in terms of 
jointly learning multiple per-snippet masks each with focus on a single 
action instance.
}
% \vspace{-0.20in}
\begin{table}[t]
	\begin{minipage}{0.5\linewidth}
		\centering
		\caption{Analysis of mask design of {\shortmodelname} on ActivityNet dataset.}
         \label{tab:maskd}
		\begin{tabular}{@{}c|cc|c@{}}
\toprule
\multirow{2}{*}{Mask Design} & \multicolumn{2}{c|}{mAP} & \multirow{2}{*}{\begin{tabular}[c]{@{}c@{}}Avg\\ masks / video\end{tabular}} \\ \cmidrule(lr){2-3}
 & 0.5 & Avg &  \\ \midrule
Actionness & 44.8 & 27.1 & 30 \\ \midrule
Our Global Mask & \textbf{56.3} & \textbf{36.5} & \textbf{250} \\ \bottomrule
\end{tabular}
	\end{minipage}\hfill
	\begin{minipage}{0.45\linewidth}
		\centering
		\captionof{figure}{Pairwise feature similarity.
% 		Cosine Similarity of a video using actionness vs our global mask
		}
		\includegraphics[height=1.4in,width=\columnwidth]{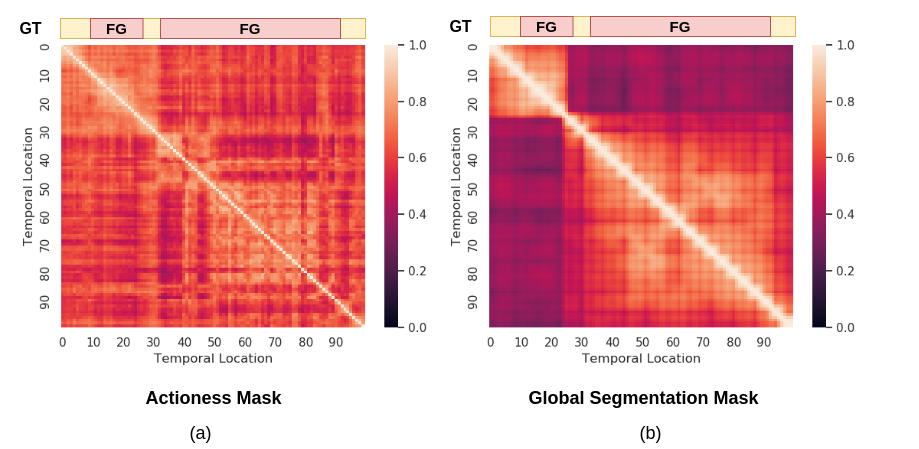}
		\label{fig:cos}
	\end{minipage}
	% \vspace{-0.30in}
\end{table}

% \noindent\textbf{Role of positional encoding}
% \textcolor{black}{We evaluate the effect of position encoding in {\shortmodelname} on ActivityNet.  
% As shown in Table \ref{tab:posenc}, it is interesting to see that position encoding is not necessary and even harmful to the performance. This indicates that with our current formulation, the snippet level temporal information does not bring extra useful information.
% }

% \begin{table}[]
%     \centering
%     \caption{Effect of positional encoding in {\shortmodelname} on ActivityNet.}
% \label{tab:posenc}
%     \renewcommand{\tabcolsep}{12pt}
%     \begin{tabular}{c|c|c}
%     \toprule
%     \multirow{2}{*}{\# Position Encoding} & \multicolumn{2}{c}{mAP} \\ \cmidrule(l){2-3} 
%      & 0.5 & Avg \\ \midrule
%     No Encoding & \textbf{56.3} & \textbf{36.5} \\
%     Learned Encoding & 53.9 & 33.4 \\
%     Fixed Encoding & 44.7 & 28.0 \\ \bottomrule
%     \end{tabular}
    
% \end{table}
\section{Limitations}
\textcolor{black}{
% Since snippet is the smallest prediction unit
In general,
short foreground and background segments with the duration similar as or less than the snippet length would challenge snippet-based TAD methods.
For instance, given short background between two foreground instances, our {\shortmodelname} might
wrongly predict it as part of the foreground. 
% that use snippets as input. 
Besides, given a snippet with mixed background and foreground,
{\shortmodelname} tends to make a background prediction.
In such cases, the ground-truth annotation involves uncertainty which however is less noted and investigated thus far.
% predicts such snippets as background.
% Also, the features generated by the off-the-shelf encoders have action and background encoded within a same snippet, our {\shortmodelname} falsely predicts such snippets as background.
% Thus, improving the sensitivity of our model on detecting short-duration action instances and extracting the foreground from noisy background will be part of future work.
}
% \begin{figure*}[!htbp]
% % \vspace{-.10in}
% \begin{center}
%   \includegraphics[scale=0.10,width=1\columnwidth]{latex/img/neurips_fig_5_v2.pdf}
% \end{center}
% % \vspace{-.15in}
% \caption{A failure case from THUMOS14
% }
% % \vspace{-.15in}
% \label{fig:failure}
% \end{figure*}

\section{Conclusion}
In this work, we have presented the first proposal-free TAD model by Global Segmentation Mask ({\shortmodelname}) learning. Instead of generating via predefined anchors, or predicting many start-end pairs (\ie, temporal proposals), our model is designed to estimate the full-video-length segmentation mask of action instances directly. As a result, the TAD model design has been significantly simplified with more efficient training and inference. With our {\shortmodelname} learning, we further show that learning global temporal context is beneficial for TAD. Extensive experiments validated that the proposed {\shortmodelname} yields new state-of-the-art performance on two TAD benchmarks, and with clear efficiency advantages on both model training and inference.

% \clearpage
\bibliographystyle{splncs04}
\bibliography{egbib}
\end{document}